\documentclass{article}

\usepackage{PRIMEarxiv}

\usepackage[utf8]{inputenc} 
\usepackage[T1]{fontenc}    
\usepackage{hyperref}       
\usepackage{url}            
\usepackage{booktabs}       
\usepackage{amsfonts}       
\usepackage{nicefrac}       
\usepackage{microtype}      
\usepackage{lipsum}
\usepackage{fancyhdr}       
\usepackage{graphicx}       
\usepackage{amssymb}
\usepackage{amsmath}
\usepackage{multirow}
\usepackage{booktabs}
\usepackage{comment}
\usepackage{float}

\title{An inpainting approach to manipulate asymmetry in pre-operative breast images}

\author{
 Helena Montenegro \\
  University of Porto \\
  INESC TEC\\
   \And
 Maria J. Cardoso \\
  University of Lisbon \\
  Fundação Champalimaud\\
  \And
 Jaime S. Cardoso \\
  University of Porto \\
  INESC TEC\\
}

\begin{document}
\maketitle

\begin{abstract}
One of the most frequent modalities of 
breast cancer treatment is surgery. Breast surgery can cause visual alterations to the breasts, due to scars and asymmetries. To enable an informed choice of treatment, the patient must be adequately informed of the aesthetic outcomes of each treatment plan. In this work, we propose an inpainting approach to manipulate breast shape and nipple position in breast images, for the purpose of predicting the aesthetic outcomes of breast cancer treatment. We perform experiments with various model architectures for the inpainting task, including invertible networks capable of manipulating breasts in the absence of ground-truth breast contour and nipple annotations. Experiments on two breast datasets show the proposed models' ability to realistically alter a patient's breasts, enabling a faithful reproduction of breast asymmetries of post-operative patients in pre-operative images.
\end{abstract}

\section{Introduction}
\label{intro}

In recent years, various surgical procedures have been developed to treat breast cancer patients, ranging from more invasive procedures that remove the entire breast (mastectomy) to conservative treatment that only removes part of the breast containing the cancer. Furthermore, patients who undergo mastectomies can also opt to reconstruct the removed breast through diverse approaches. These treatments can cause substantial alterations to a patient's breasts, such as visible scars and perceived asymmetries. Furthermore, the aesthetic outcomes of breast cancer treatment can vary widely depending on various factors related to the patient, the tumour and the type of treatment.

The choice of treatment plan is often challenging for breast cancer patients, due to a lack of awareness of potential aesthetic results and the difficulty in visualizing the effects of treatment in their breasts. Furthermore, unrealistic expectations often lead patients to be disappointed with the treatment results. 
As the aesthetic outcome of surgery may have a long-lasting impact on the patients' self-esteem, the patients must be adequately informed about the potential aesthetic results of each surgical procedure when deciding on the treatment plan. This work aims to help patients calibrate expectations and select the most fitting treatment plan by showing them a prediction of their own possible aesthetic outcome.

Freitas \textit{et al.} \cite{freitas2024isbi} and Montenegro \textit{et al.} \cite{montenegro2024aim} introduced the first efforts towards predicting the aesthetic outcomes of breast cancer treatment by proposing three methods to reproduce the asymmetries of post-operative patients on pre-operative images.  
However, the methods were either incapable of realistically altering the nipple or produced low-quality images that did not resemble the patient. In this work, we propose an inpainting approach to manipulate breast shape and nipple position in breast images, enabling the reproduction of asymmetries caused by treatment on a pre-operative patient. In specific, we cover the breast submitted to surgery with a segmentation mask and annotate the desired position of the nipple and lower breast contour within the mask. Then, we develop inpainting networks that inpaint the missing breast so that its shape and nipple match the annotations. We experiment with various autoencoder-based architectures for inpainting, proposing a new strategy that takes advantage of the natural symmetry of breast images to manipulate the breast while retaining characteristics such as color progression. Furthermore, we propose invertible networks to simultaneously inpaint and segment the breasts, removing the need to manually annotate the breast contour. Experiments on private and public \cite{guo2022fully} breast datasets show the inpainting models' capacity to manipulate asymmetries in breast images. 

To summarize, the main contributions of this work are:
\begin{itemize}
    \item Proposal of symmetry-based inpainting strategy for breast images;
    \item Proposal of invertible inpainting strategies for automatic nipple and lower breast contour detection and inpainting on breast images;
    \item Evaluation of the proposed inpainting at reconstructing and manipulating breast images using two breast datasets;
    \item Evaluation of best-performing approaches at transferring asymmetries from post-operative breast images into pre-operative images, to predict the aesthetic outcomes of breast cancer treatment.
\end{itemize}

\section{Related Work}

The following sections present related work on the topics of aesthetic outcomes of breast cancer treatment, inpainting and invertible networks.

\subsection{Aesthetic Outcomes of Breast Cancer Treatment} \label{sec:related-work}

The aesthetic evaluation of breast cancer treatment has been extensively researched in the literature, mostly focusing on objective metrics to quantify breast asymmetries. The existing metrics were proposed to evaluate two aspects of asymmetries within the breasts: nipple displacement and differences in the shape of the breasts. Regarding nipple asymmetries, Breast Retraction Assessment (BRA) \cite{pezner1985breast} quantifies the horizontal and vertical displacement of the nipples in relation to the sternal notch. Breast Compliance Evaluation (BCE) \cite{tsouskas1990breast} evaluates the vertical displacement of the nipple in relation to the lowest point of the breast contour. Horizontal Nipple Retraction (HNR) \cite{freitas2024isbi} measures the distance from each nipple to the external contour of the breast. Upwards Nipple Retraction (UNR) \cite{van1989cosmetic} measures the vertical distance between the nipples. Regarding metrics to measure asymmetries in the shape of the breasts, Lower Breast Contour (LBC) \cite{van1989cosmetic}  measures the vertical distance between the lowest point of each breast. Breast Contour Difference (BCD), Breast Area Difference (BAD) and Breast Overlap Difference (BOD) \cite{cardoso2007towards} measure the differences between the contour and shape of the breasts. These last three metrics are highly dependent on the placement of the endpoints of the breast contour, which is often characterized by high variability as it is difficult to pinpoint where the breast ends in a frontal image. In this work, we use these metrics to evaluate the proposed methodology. 

Regarding the prediction of the aesthetic results of breast cancer treatment, Freitas \textit{et al.} \cite{freitas2024isbi} propose two approaches to transfer breast asymmetries from post-operative patients into pre-operative images. First, they apply image warping to reposition the breast's contour. Nevertheless, this approach is incapable of manipulating the nipple and requires the manual annotation of the breast contour during inference. On a second approach, the authors use a weakly-supervised learning approach based on disentangled representation learning to isolate asymmetries in breast images from the patient's identity, enabling to alter asymmetries while preserving identity-related traits. Although this method succeeds at altering both breast and nipple, it overly changes the shape of the pre-operative patients' breasts to match those of the post-operative patients. Montenegro \textit{et al.} \cite{montenegro2024aim} extend the weakly-supervised approach by implementing a symmetry-based loss to promote higher resemblance between the original and generated images. Nevertheless, the resulting images lack visual quality. In this work, we propose a novel method to manipulate breast images that is simultaneously capable of generating realistic images and manipulating both the nipple and breast, through conditional image inpainting. 

\subsection{Inpainting}

Image inpainting corresponds to the image-to-image translation task of filling missing regions in images in coherence with the surrounding content. Inpainting has been applied for various purposes, including object removal and partial image generation. Most works train networks such as Autoencoders (AE), U-Net \cite{ronneberger2015u}, and Fully Convolutional Networks \cite{long2015fully} to fill missing regions by reconstructing the ground-truth image \cite{liu2018image, xiang2023deep}. More recently, various works take advantage of deep generative models like Generative Adversarial Networks (GANs) \cite{goodfellow2014gans} or Denoising Diffusion Probabilistic Models (DDPMs) \cite{ho2020denoising} to achieve more realistic predictions and higher variability within the inpainted region \cite{zeng2021cr, pathak2016context, lim2023image, corneanu2024latentpaint}. Within inpainting, there are two research lines that are particularly relevant for this work: conditional inpainting, where the inpainting process is conditioned on some variable, and applications of inpainting to medical imaging. 

Regarding conditional inpainting, works on face inpainting often condition the inpainting network on facial landmarks by concatenating them with the input image that contains the missing regions \cite{sun2018natural, yang2020generative, hong2024conditional}. Other works condition the inpainting process on reference images \cite{lim2023image} or on textual cues \cite{xie2023smartbrush} by providing the respective embeddings as input to the network. In our case, we condition the inpainting process on nipple and lower breast annotations drawn over the covered breast, enabling the network to automatically understand where the nipple and lower breast should be drawn, and minimizing the amount of annotations required for the inpainting process. 

Within medical imaging, various works use GANs and autoencoders to improve the visual quality of the images by inpainting square regions, stripes or other arbitrary regions within medical images such as magnetic resonance images \cite{armanious2019adversarial, chai2020mri, armanious2020ipa} and chest X-rays \cite{tran2020deep}. Other works use inpainting to manipulate medical images for purposes such as straightening a patient's spine \cite{bukas2021patient}. Nevertheless, in all these works, the information that is covered in the input image, which may be useful for a more faithful reconstruction of the image, is not provided to the model and is, therefore, lost. In this work, we propose an inpainting model that uses information from the covered breast region to inpaint it, enabling a more realistic reconstruction and manipulation of the breast. 

\subsection{Invertible Networks}

Invertible networks perform exclusively invertible operations on their inputs, making it possible to obtain an input based on its output. These networks are widely used as the building blocks of normalizing flows \cite{rezende2015variational} for unsupervised image generation tasks. For instance, RealNVP \cite{dinh2017density} uses affine coupling layers that apply invertible affine transformations to the data. Glow \cite{kingma2018glow} uses invertible $1\times 1$ convolutions along with affine coupling layers. More recently, invertible architectures, such as i-RevNet \cite{jacobsen2018irevnet}, have been used for image classification tasks, enabling the reconstruction of the original image based on its latent features and avoiding loss of information. 
More similarly to our work, Liu \textit{et al.} \cite{liu2020deep} develop an invertible autoencoder to perform image-to-image translation tasks like inpainting, by employing two Glow models as the encoder and decoder of the network. However, the autoencoder is only trained and evaluated in the inpainting direction, discarding the inverse direction of removing regions from an image. In our work, we develop invertible autoencoders based on Glow and i-RevNet, training them to perform both inpainting and segmentation on breast images, enabling the manipulation of unlabelled breast images.

\section{Methodology}

\begin{figure}[h]
\centering
\includegraphics[width=1.0\textwidth]{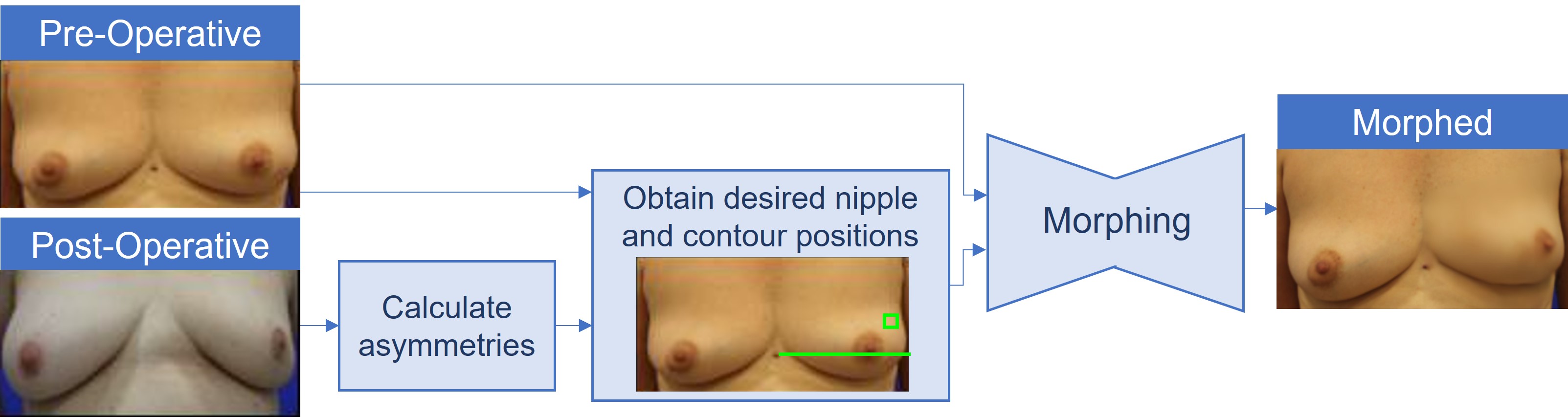}
\caption{Generic pipeline to transfer breast asymmetries from post-operative patients into pre-operative images.}
\label{fig:transferring-asymmetries-generic-overview}
\end{figure}

This work aims to predict the potential outcomes of breast cancer treatment based on a pre-operative breast image of the patient.  Since most available datasets do not contain pairs of pre- and post-surgery images of past patients, the model must learn traits caused by treatment from mostly post-operative images. As such, we design an approach to transfer asymmetries caused by treatment from post-operative past patients into a pre-operative image of the current patient proposed to be submitted to an identical operation. More specifically, given a pre-operative image and annotations of the desired position of the nipple and contour of the breast (defined according to the asymmetries of the post-operative image), the model should predict a morphed image whose nipple and breast match the annotations, as depicted in Fig.~\ref{fig:transferring-asymmetries-generic-overview}. 

Since there is no ground-truth describing how the pre-operative images would look like with the asymmetries of post-operative images, training a model to directly predict the morphed images is difficult. To overcome this limitation, we design strategies to train a model to realistically inpaint a breast image with a missing breast annotated with the desired positions of the nipple and lowest breast contour. The following sections describe the proposed inpainting techniques to reconstruct and manipulate breast images, and the approach to transfer asymmetries from post-operative into pre-operative images using the proposed inpainting models.

\subsection{Inpainting}

\begin{figure}[h]
\centering
\includegraphics[width=1.0\textwidth]{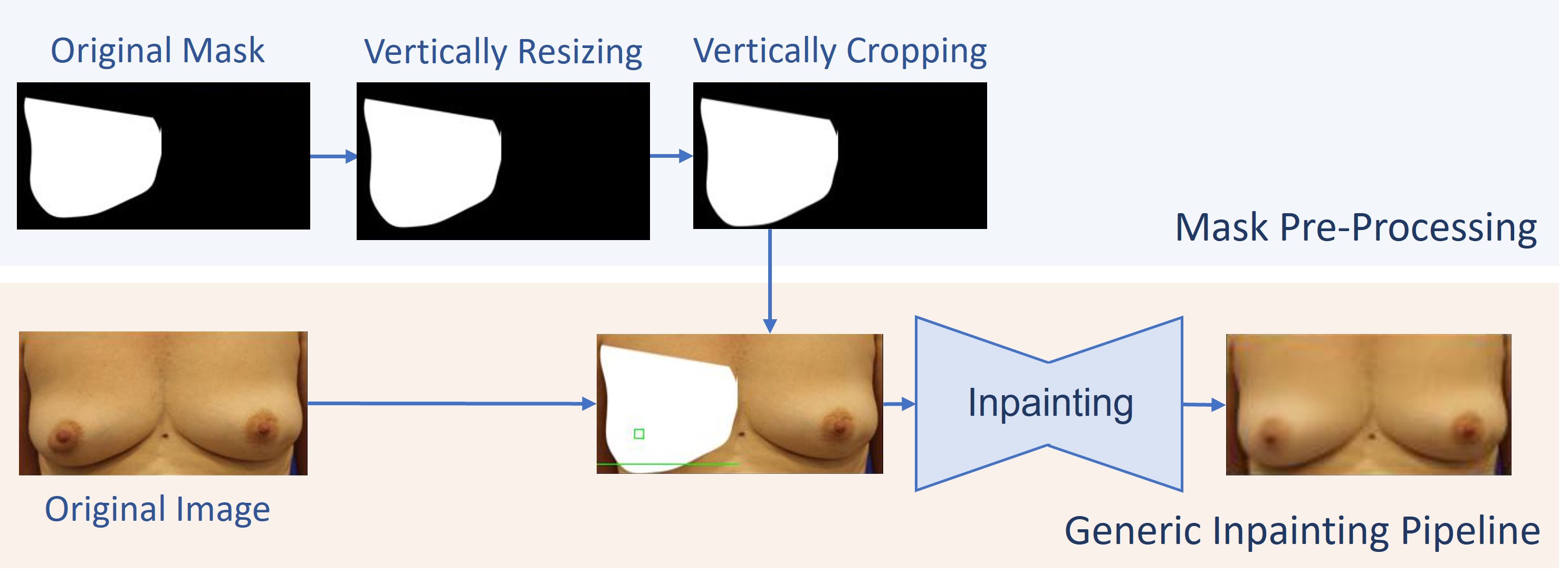}
\caption{Generic overview of inpainting framework.}
\label{fig:inpainting-overview}
\end{figure}

Given an image of the two breasts, annotated with keypoints delineating their contour and the position of the nipples, we cover one of the breasts using its segmentation mask. Since we want to manipulate not only the inside of the breast but also its shape, we stretch the mask so that it reaches the bottom of the image, by first vertically resizing it and then cropping it according to the original image size, as shown in Fig.~\ref{fig:inpainting-overview}. Then, we draw a square over the mask, indicating the position of the nipple, and a horizontal line intersecting the lowest point of the breast contour. Finally, we train an inpainting model on the masked image to inpaint the missing region, by minimizing the mean squared error between the resulting image and the original. Since the network is mostly trained with asymmetric post-operative images with differently shaped breasts, it learns to associate the nipple and lowest breast contour to the provided square and line annotations. As a result, by repositioning the square and the horizontal line, we can alter the nipple and breast shape. 

\begin{figure}[h]
\centering
\includegraphics[width=1.0\textwidth]{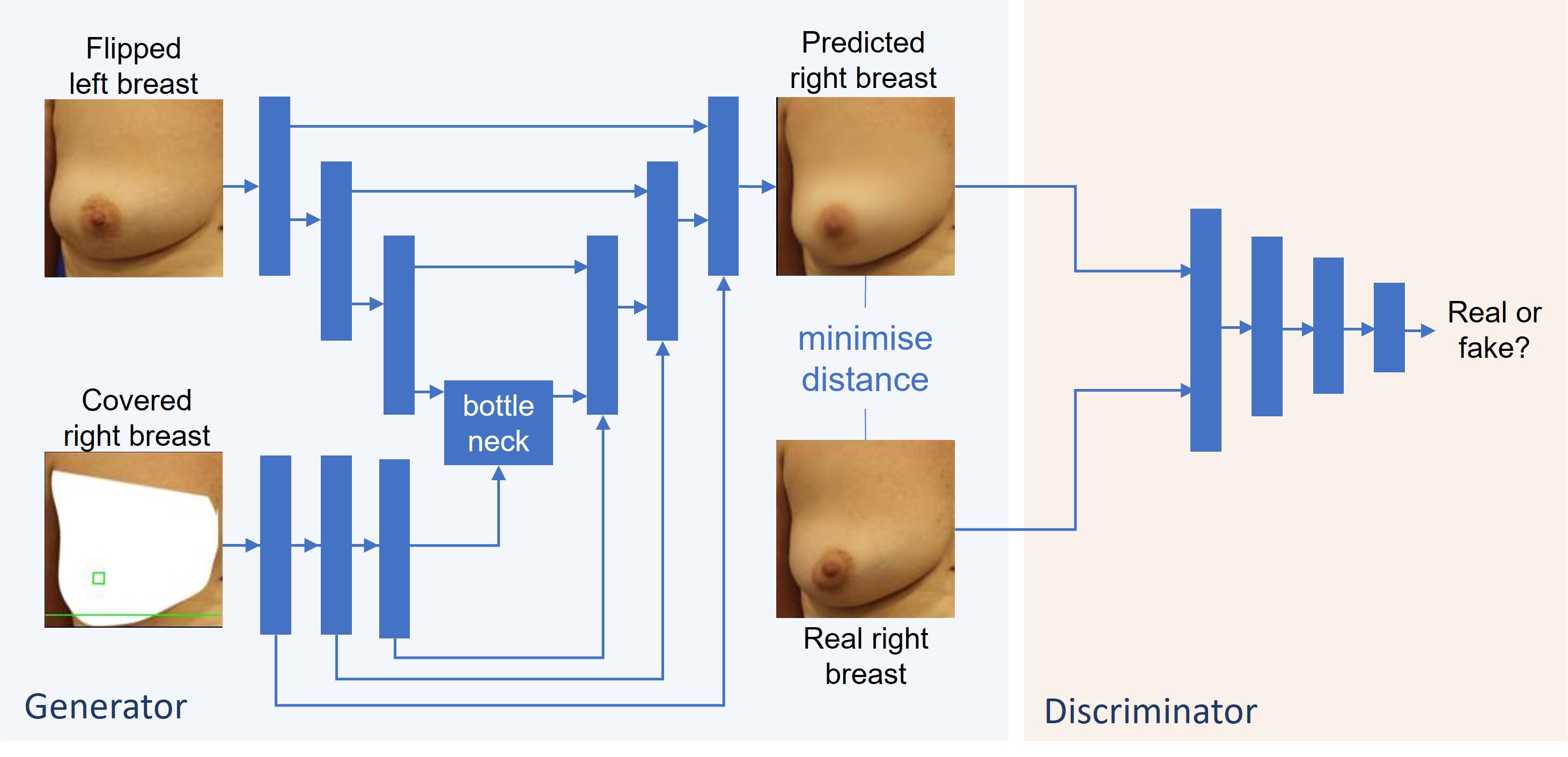}
\caption{Proposed Double GAN model. During training the model receives the masked right breast and a flipped version of the left breast, and is trained to predict the right breast. On inference, it receives the masked and original versions of the right breast to inpaint the breast while preserving its color progression.}
\label{fig:proposed-unet}
\end{figure}

In order to realistically inpaint the breast, the model should obtain information regarding the structure of the nipple and color progression within the breast from the unmasked breast. Nevertheless, there is no guarantee that the original covered breast looks identical to the uncovered one. As such, covering the original breast leads to a substantial loss of information. To target this limitation, we propose an inpainting model based on U-Net that uses the information of the original breast while inpainting it. More specifically, the proposed model is a U-Net with two encoders that receives an image of the isolated breast and a masked breast image, and predicts the inpainted image. We train this U-Net (Double GAN) as the generator of a GAN, using adversarial training to enhance the realism of the predicted images, as illustrated in Fig.~\ref{fig:proposed-unet}. During training, the U-Net receives the horizontally flipped version of the left breast and a masked image of the right breast, and learns to reconstruct the right breast. During inference, the model receives both the right breast and its covered version with re-positioned nipple and breast annotations, and it produces the manipulated version of the right breast. It is possible to manipulate the left breast by horizontally flipping the image before applying the network.

We compare the proposed network with a typical U-Net (Simple U-Net) that receives the entire image with the covered breast and predicts the original image. In terms of architecture, the encoder and decoder of the Simple U-Net are constituted by four blocks of convolutional layers each. Furthermore, we also compare these networks with an Attention U-Net that introduces self-attention layers into the encoder and decoder \cite{vaswani2017attention}. 


During the training of all these models, we use two data augmentation strategies that take advantage of the natural symmetry of breast images: horizontal flips and mirroring the images based on the vertical axis positioned at the center of the image. Nevertheless, to avoid the networks ignoring the annotated nipple position and breast lower contour, we only mirror the images with 10\% probability.

\subsection{Invertible Inpainting}

\begin{figure}[h]
\centering
\includegraphics[width=1.0\textwidth]{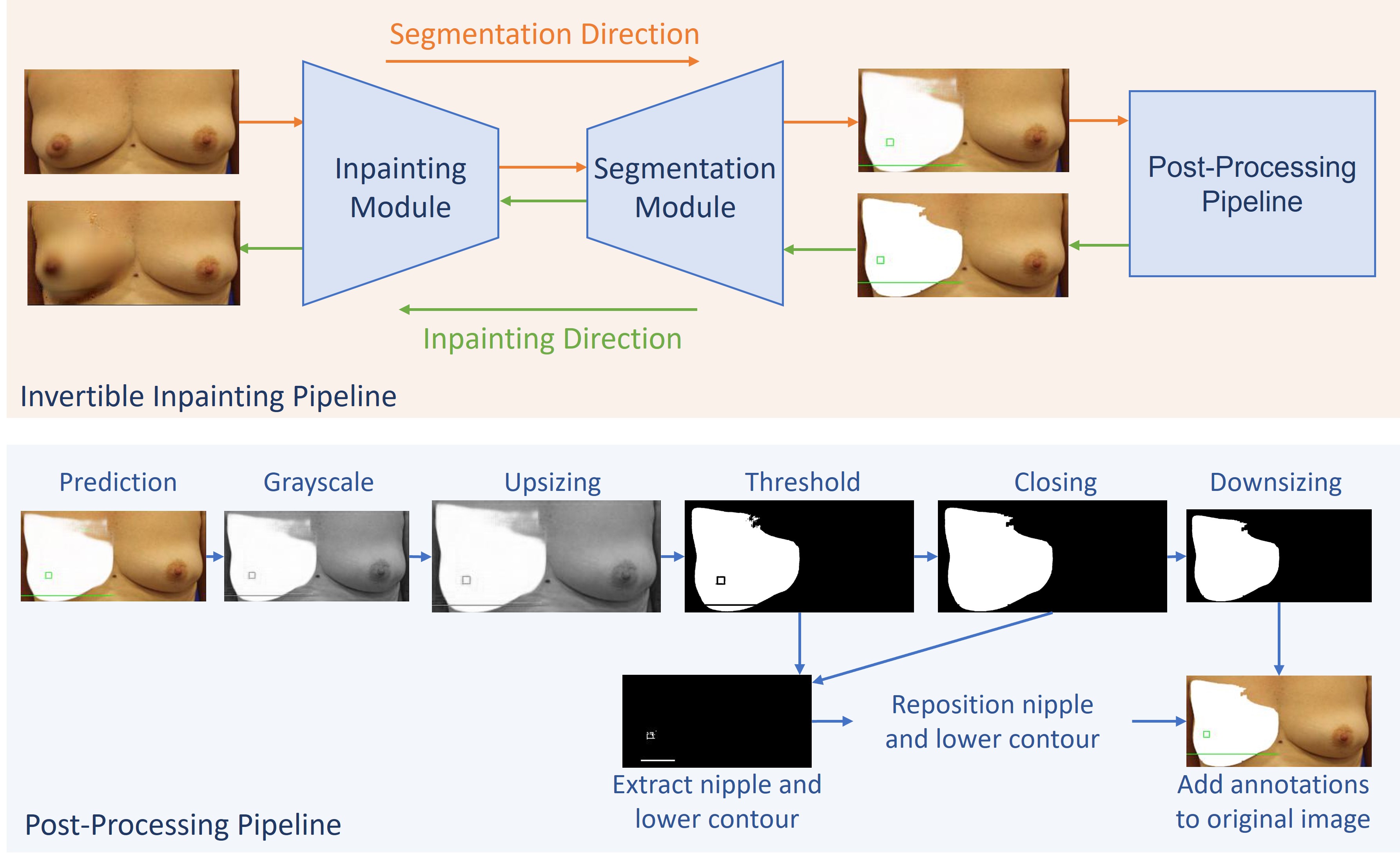}
\caption{Overview of invertible inpainting approach for manipulating breast asymmetries without requiring annotations. During training, the segmentation direction receives the original image and predicts its masked version, and the inpainting direction receives the masked image and predicts the original. During inference, the original image goes through the segmentation direction to obtain its masked version. Then, the post-processing pipeline improves the quality of the masked image and repositions the nipple and breast. Finally, the inpainting direction obtains the morphed image.}
\label{fig:invertible-overview}
\end{figure}

The model proposed in the previous section requires annotations of the breast contour and nipple position. However, annotating the images requires effort from clinicians, making it more difficult to adopt such techniques in clinical practice. As such, we explore invertible networks that simultaneously segment and inpaint the breast so that the breasts can be manipulated without requiring manual annotations. In specific, we replace the encoder and decoder of a typical U-Net with invertible networks, which we call the inpainting module and the segmentation module, as shown in Fig.~\ref{fig:invertible-overview}.  As such, the proposed network can receive an unmasked breast image and output its masked and annotated version, through the segmentation direction. Furthermore, it can receive a masked breast image and output the inpainted image through the inpainting direction. We train the network in both directions, using the breast image and its masked version as both the input and ground-truth, and using mean squared error as the reconstruction loss in each direction. In these networks, we only mask the right breast to ensure that the segmentation network always segments the same breast while preserving the other. As such, to segment or manipulate the left breast, we horizontally flip the image before applying the network.
For the inpainting and segmentation modules, we experiment with two of the most well-known invertible networks: Glow \cite{kingma2018glow} and i-RevNet \cite{jacobsen2018irevnet}.

The invertible inpainting network can be applied to an unlabelled breast image by segmenting it through the segmentation direction, obtaining the breast segmentation mask and the positions of the nipple and lower breast, covering the breast of the original image with the segmentation mask with repositioned nipple and lower breast annotations, and applying the inpainting direction to obtain the manipulated image. To obtain the breast segmentation mask based on the predicted masked breast, we use the post-processing pipeline illustrated in Fig.~\ref{fig:invertible-overview}. More specifically, we threshold the grayscale masked image to obtain only the white mask that covers the breast, and apply the morphological operation closing, corresponding to a dilation followed by an erosion, to remove the square and line annotations. To avoid obtaining pixelated segmentation masks, we resize the masked image to 5 times its size before applying the pipeline and then resize it back to the original size afterwards. 

To obtain the predicted nipple and breast positions in the segmented images, we first subtract the mask obtained after the closing operation from the mask before this operation, obtaining a mask that only contains the square representative of the nipple and the horizontal line indicating the lower breast contour. Then, we search the image for the horizontal line with the highest amount of white pixels, corresponding to the lower breast. To obtain the coordinates of the nipple, we search for the rightmost column with a number of white pixels higher than a threshold $t$ (in our case, $t = 7$), chosen according to the size of the square (in our case, 9x9), corresponding to the rightmost edge of the square. Finally, we consider the y-coordinate as the mean of the vertical position of the white pixels in the column and obtain the x-coordinate by subtracting half the edge of the square (in our case, 4 pixels) from the column. 

\subsection{Transferring Asymmetries between Breast Images}

\begin{figure}[h]
\centering
\includegraphics[width=1.0\textwidth]{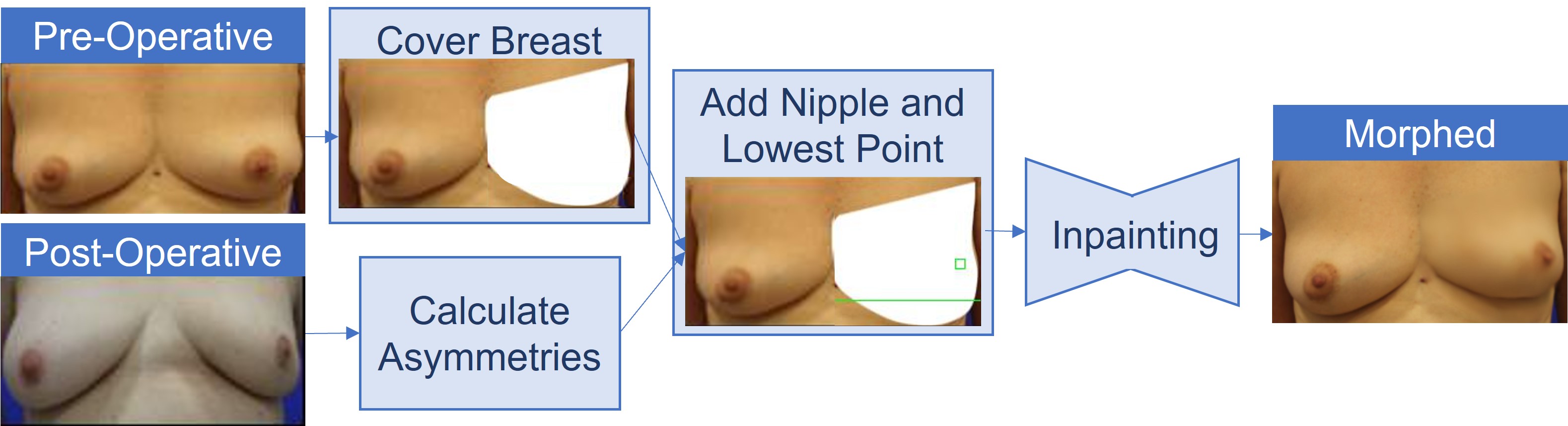}
\caption{Pipeline to transfer breast asymmetries from post-operative patients into pre-operative images using inpainting networks.}
\label{fig:transferring-asymmetries-overview}
\end{figure}

After obtaining networks capable of manipulating the breast shape and nipple position, we can apply these networks to transfer asymmetries from post-operative patients into pre-operative breast images according to the pipeline illustrated in Fig.~\ref{fig:transferring-asymmetries-overview}. We use the asymmetries of post-operative images to calculate the new position of the nipple and lower breast contour. 

We calculate the new lower breast position of the target breast $y_{tar}$ of the pre-operative patient so that its relative position to the lower contour of the other breast $y_{oth}$ matches that of the post-operative patient ($y'_{tar}$ and $y'_{oth}$), as depicted in Eq.~\ref{eq:lbc-asymmetry}. 

\begin{eqnarray}
\begin{aligned}
\frac{y_{tar}}{y_{oth}} = \frac{y'_{tar}}{y'_{oth}} \Leftrightarrow  y_{tar} = y_{oth} \frac{y'_{tar}}{y'_{oth}} \label{eq:lbc-asymmetry}
\end{aligned}
\end{eqnarray}

We also apply Eq.~\ref{eq:lbc-asymmetry} to vertically reposition the nipple, where $y$ and $y'$ now refer to the vertical position of the nipple. 
Finally, we calculate the horizontal position of the nipple so that the relative distances between the nipple and the external contour of the breasts match those of the post-operative patient. To do so, we calculate the relative distance from the nipple $x_{nip}$ to the external contour $x_{ext}$ normalized by the distance between the external and internal $x_{int}$ contours according to Eq.~\ref{eq:hnr-relative-distance}.

\begin{eqnarray}
\begin{aligned}
\mathcal{D}(x) = \frac{x_{ext} - x_{nip}}{x_{ext}-x_{int}} \label{eq:hnr-relative-distance}
\end{aligned}
\end{eqnarray}

Then, we calculate the horizontal position of the target nipple of the pre-operative patient $x_{tar}$ based on the position of the other nipple $x_{oth}$ and on the corresponding nipples of the post-operative patient $x'_{tar}$ and $x'_{oth}$, according to Eq.~\ref{eq:hnr-asymmetry}.

\begin{eqnarray}
\begin{aligned}
\frac{\mathcal{D}(x_{tar})}{\mathcal{D}(x_{oth})} = \frac{\mathcal{D}(x'_{tar})}{\mathcal{D}(x'_{oth})} \Leftrightarrow \frac{x_{ext} - x_{nip}}{x_{ext}-x_{int}} = \mathcal{D}(x_{oth}) \frac{\mathcal{D}(x'_{tar})}{\mathcal{D}(x'_{oth})} \Leftrightarrow \\
\Leftrightarrow x_{nip} = x_{ext} - (x_{ext} - x_{int}) \mathcal{D}(x_{oth}) \frac{\mathcal{D}(x'_{tar})}{\mathcal{D}(x'_{oth})}
\label{eq:hnr-asymmetry}
\end{aligned}
\end{eqnarray}

\section{Experiments}

In the experiments, we use a private dataset of 2,000 pre- and post-operative breast images acquired by the Portuguese Champalimaud Foundation as part of the CINDERELLA Project\footnote{https://cinderellaproject.eu/project/} (CP) \cite{kaidar2023evaluating}. These images were annotated by breast surgeons with keypoints that delineate the breast contour and nipple position, and aesthetic evaluation labels that classify the aesthetic results as Poor, Fair, Good or Excellent. We use 150 images for testing purposes and the remaining for training. Out of the 150 images used for testing, 50 are pre-operative images, 50 are post-operative images that were considered Excellent or Good by clinicians, and 50 represent Fair and Poor post-operative images with bigger asymmetries. 
As a second dataset, we use the publicly available Breast Cosmetic (BC) dataset \cite{guo2022fully} of 3,762 post-operative images annotated with breast contour and nipple position. In this case, we use 10\% of the images for testing and the remaining for training. All images in both datasets were resized to 128x256 in all experiments, except with the attention U-Net, whose architecture requires higher computational resources and thus was trained on smaller images of dimensions 64x128.

The following sections expose the results of the inpainting networks in reconstructing and manipulating the original images, as well as the results of transferring asymmetries from post-operative patients into pre-operative images.

\subsection{Inpainting Results}

In a first experiment, we compare the results of the proposed inpainting networks in reconstructing the original images based on masked images with nipple and breast contour annotations in their original positions. In addition to the proposed models, we present the results of training only the generator of the Double GAN without the adversarial discriminator (Double U-Net) and without the skip connections between the second encoder and the decoder (Conditional U-Net). All networks were trained for 1,000 epochs. As evaluation metrics, we use Structural Similarity Index Measure (SSIM) \cite{wang2004ssim}, Peak Signal-to-Noise Ratio (PSNR) and Learned Perceptual Image Patch Similarity (LPIPS) \cite{zhang2018unreasonable}. The results are depicted in Table~\ref{tab:reconstruction-results}. In the invertible networks, we expose the results obtained by applying only the inpainting direction, considering fully annotated (FA) images, and both the inpainting and segmentation directions, considering not annotated (NA) images.

\begin{table}[h]
\centering
\begin{tabular}{l|ccc|ccc}
\toprule
\multirow{2}{*}{Method} & \multicolumn{3}{c|}{CP} & \multicolumn{3}{c}{BC} \\
& SSIM & LPIPS & PSNR & SSIM & LPIPS & PSNR \\ \midrule
Simple U-Net & 0.935 & 0.137 & 30.38 & 0.957 & 0.087 & 32.49 \\
Attention U-Net & 0.950 & 0.081 & 32.76 & 0.966 & 0.053 & 34.05 \\
Conditional U-Net & 0.932 & 0.159 & 30.48 & 0.936 & 0.149 & 31.46 \\
Double U-Net & 0.948 & 0.123 & 32.26 & 0.961 & 0.079 & 33.54 \\
Double GAN & \textbf{0.971} & \textbf{0.060} & \textbf{36.29} & \textbf{0.972} & \textbf{0.052} & \textbf{35.94} \\
i-RevNet AE (FA) & 0.826 & 0.234 & 28.82 & 0.907 & 0.149 & 30.53 \\
i-RevNet AE (NA) & 0.834 & 0.236 & 28.78 & 0.869 & 0.201 & 29.22 \\
Glow AE (FA) & 0.901 & 0.201 & 29.94 & 0.945 & 0.098 & 32.06 \\
Glow AE (NA) & 0.890 & 0.212 & 29.16 & 0.932 & 0.117 & 29.90 \\
\bottomrule
\end{tabular}
\caption{Results of SSIM $\uparrow$, LPIPS $\downarrow$ and PSNR $\uparrow$ inpainting for image reconstruction. }
\label{tab:reconstruction-results}
\end{table}

\begin{figure}[h]
\centering
\includegraphics[width=1.0\textwidth]{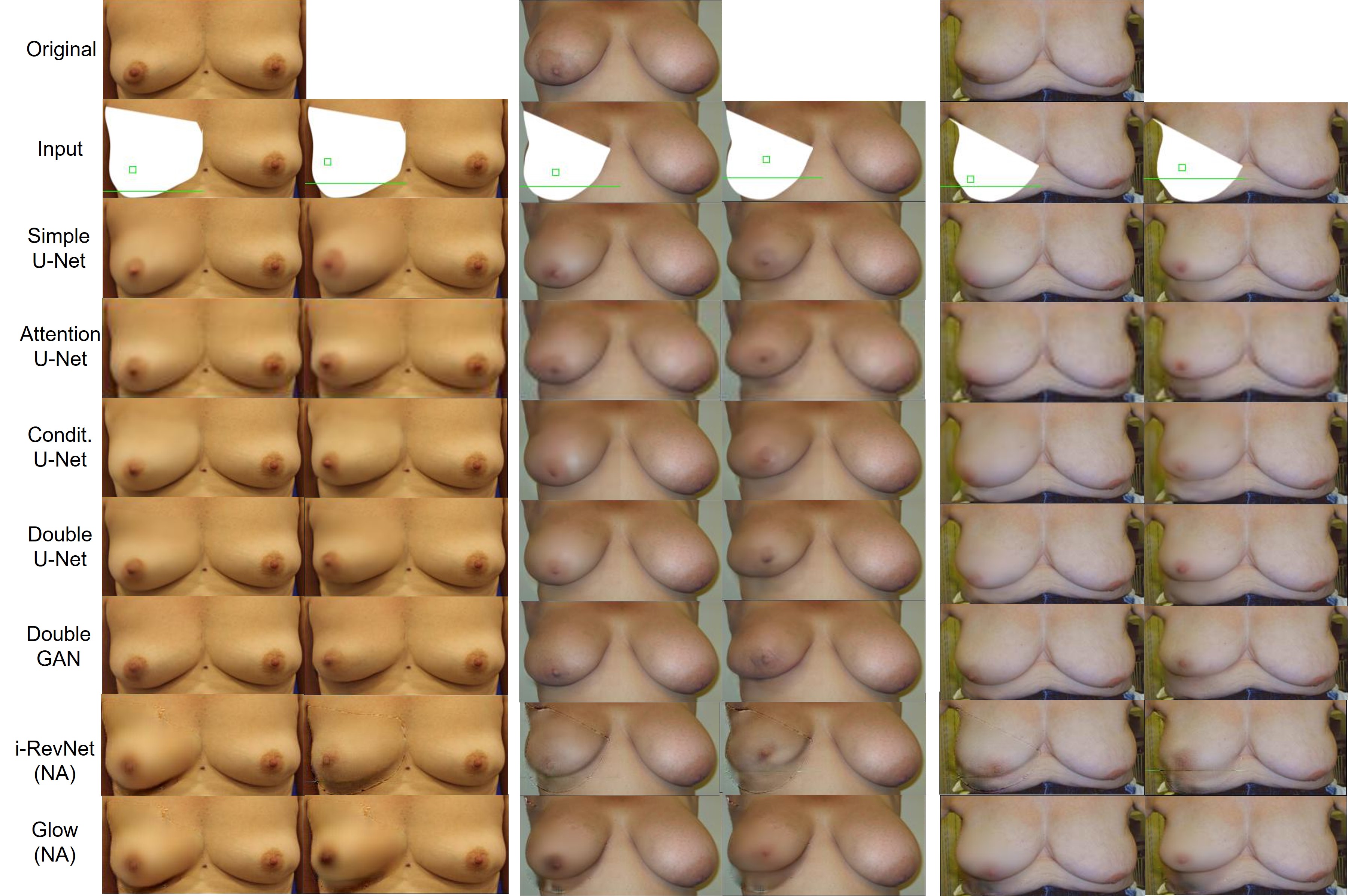}
\caption{Results of reconstruction and manipulation using the inpainting networks on the CP (first column) and BC (second and third columns) datasets.}
\label{fig:inpainting-results}
\end{figure}

We observe that the network with higher capacity to reconstruct the original image on both datasets is the proposed Double GAN, followed by the attention U-Net. In general, the invertible networks achieve the worst results, with the Glow-based autoencoder surpassing i-RevNet. Nevertheless, unlike the non-invertible models, the invertible networks are capable of manipulating unlabeled images with a small loss of realism when compared with their application on annotated images. Fig.~\ref{fig:inpainting-results} shows examples of results where it is possible to see that the networks are capable of successfully manipulating the lower breast and nipple position.

\subsection{Segmentation Results with Invertible Networks}

On Table~\ref{tab:segmentation-results}, we evaluate the segmentation direction of the inpainting networks by comparing the predicted mask with the ground-truth, using Intersection over Union (IoU). Furthermore, we present the mean absolute error between the ground-truth annotations and the predicted lower breast contour (LB), and horizontal (NX) and vertical (NY) position of the nipple.

\begin{table}[H]
\centering
\begin{tabular}{llcccc}
\toprule
Data  & Method & IoU $\uparrow$ & LB $\downarrow$ & NX $\downarrow$ & NY $\downarrow$ \\ \midrule
\multirow{2}{*}{CP} 
& i-RevNet & \textbf{0.867} & 6.120 & 11.333 & 9.013  \\
& Glow & 0.858 & \textbf{1.693} & \textbf{4.280} & \textbf{4.253}  \\
\midrule
\multirow{2}{*}{BC} 
& i-RevNet & 0.841 & 5.753 & 16.209 & 10.802  \\
& Glow & \textbf{0.888} & \textbf{0.995} & \textbf{3.842} & \textbf{3.430} \\
\bottomrule
\end{tabular}
\caption{Results of segmentation and breast annotation.}
\label{tab:segmentation-results}
\end{table}

\begin{figure}[h]
\centering
\includegraphics[width=0.8\textwidth]{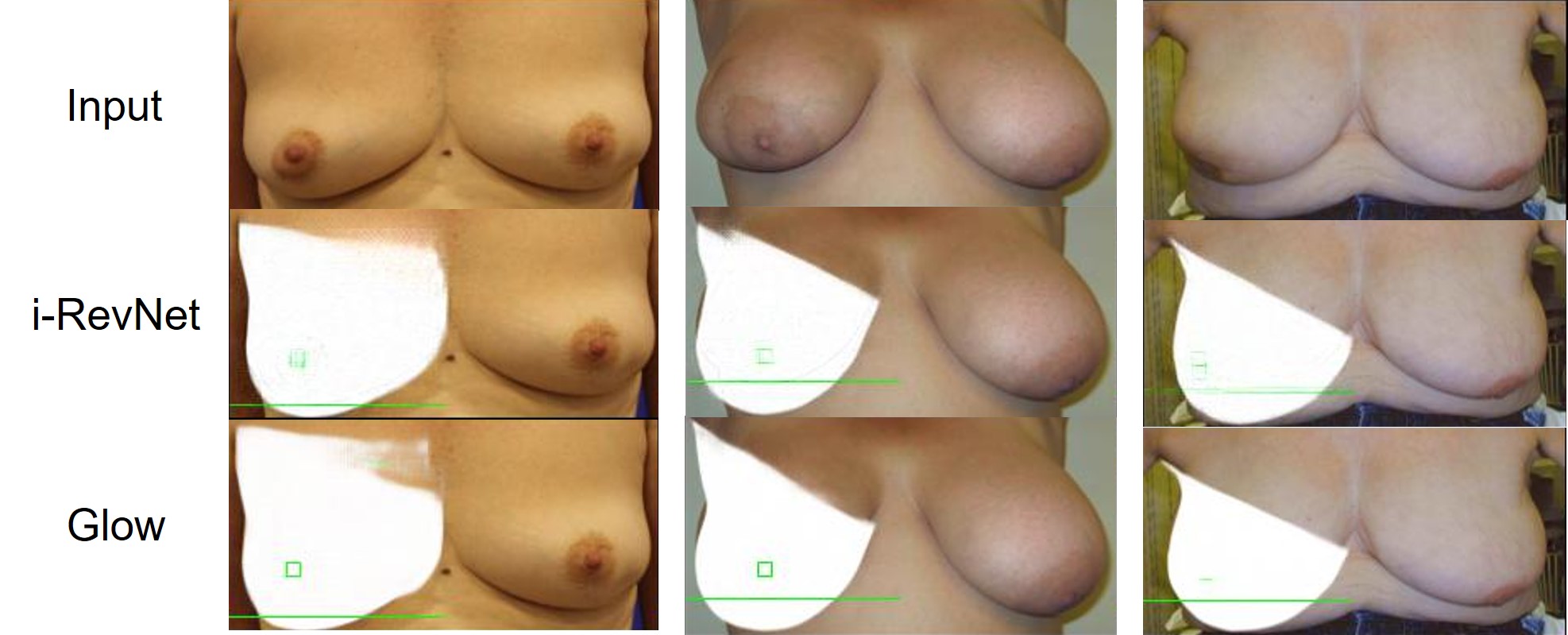}
\caption{Examples of segmentation results on the CP and BC datasets.}
\label{fig:segmentation-results}
\end{figure}

The Glow-based model produces more accurate predictions of lower breast contour and nipple annotations for both datasets. Both networks are capable of predicting a mask to cover the breast, achieving relatively high IoU on both datasets. Fig.~\ref{fig:segmentation-results} shows examples of results of the segmentation process.

\subsection{Transferring Asymmetries}

In this section, we apply the proposed inpainting approaches that achieved the best results at image inpainting to transfer asymmetries from post-operative images into pre-operative ones on the CP dataset. In specific, we compare the GAN-based U-Net, the attention U-Net and the Glow-based model with the state of the art. To do so, we generate 100 images for each network by transferring the asymmetries of the post-operative test images into the pre-operative ones. Then, we manually annotate the resulting images with keypoints delineating the breast contour and nipple position and calculate the asymmetry metrics introduced in Section~\ref{sec:related-work}. Finally, we calculate the mean absolute error between the asymmetry metrics of the generated image and the template post-operative image, presenting the results in Table~\ref{tab:final-results}. Furthermore, we use SSIM to measure the similarity between the original and manipulated images. In the Glow-based model, we compare its application to fully annotated images, not annotated pre-operative images with annotated templates, and without any annotations. 

\begin{table}[t!]
\centering
\begin{tabular}{l|l|cccc|cccc|c}
\toprule
\multirow{2}{*}{T} & \multirow{2}{*}{Method} & \multicolumn{4}{c|}{Nipple Asymmetries} & \multicolumn{4}{c|}{Breast Asymmetries} &\multirow{2}{*}{SSIM} \\
 & & BRA & UNR & BCE & HNR & LBC & BCD & BAD & BOD & \\ 
\midrule 
& Baseline & 0.044 & 0.041 & 0.105 & 0.075 & 0.026 & 0.032 & 0.070 & 0.115 & 0.584\\ 
& CV \cite{freitas2024isbi} & 0.045 & 0.034 & 0.137 & 0.050 & \textbf{0.001} & \textbf{0.023} & 0.061 & 0.092 & 0.898\\
G & A. U-Net & \textbf{0.042} & 0.023 & 0.106 & 0.030 & 0.016 & 0.037 & 0.068 & 0.096 & 0.904 \\ 
O& D. GAN & 0.043 & 0.025 & 0.105 & 0.032 & 0.012 & 0.035 & 0.072 & 0.096 & \textbf{0.930} \\ 
O& Glow (FA) & 0.044 & \textbf{0.021} & \textbf{0.085} & 0.032 & 0.015 & 0.034 & 0.065 & 0.105 & 0.868 \\  \cmidrule{2-11}
D& DL \cite{freitas2024isbi} & 0.044 & 0.025 & 0.092 & 0.033 & 0.010 & 0.026 & \textbf{0.046} & 0.106 & 0.635 \\ 
& SDL \cite{montenegro2024aim} & 0.043 & 0.024 & 0.086 & \textbf{0.028} & 0.011 & 0.031 & 0.051 & \textbf{0.090} & 0.655 \\ 
& Glow (AT) & 0.052 & 0.041 & 0.137 & 0.062 & 0.015 & 0.036 & 0.075 & 0.110 & 0.865 \\
& Glow (NA) & 0.054 & 0.045 & 0.141 & 0.057 & 0.017 & 0.049 & 0.096 & 0.095 & 0.862 \\ 
\midrule
& Baseline & 0.090 & 0.108 & 0.181 & 0.081 & 0.080 & 0.095 & 0.158 & 0.153 & 0.564\\ 
& CV \cite{freitas2024isbi} & 0.056 & 0.060 & 0.193 & 0.056 & \textbf{0.002} & 0.052 & 0.102 & 0.091 & 0.884\\
P & A. U-Net & 0.054 & 0.048 & 0.179 & \textbf{0.042} & 0.022 & 0.050 & 0.099 & 0.090 & 0.877 \\ 
O& D. GAN & \textbf{0.047} & \textbf{0.045} & \textbf{0.156} & 0.049 & 0.019 & 0.053 & 0.104 & \textbf{0.088} & \textbf{0.914} \\ 
O& Glow (FA) & 0.052 & 0.047 & 0.156 & 0.043 & 0.029 & 0.056 & 0.110 & 0.106 & 0.854 \\ \cmidrule{2-11}
R& DL \cite{freitas2024isbi} & 0.051 & 0.058 & 0.170 & 0.063 & 0.025 & \textbf{0.039} & \textbf{0.081} & 0.089 & 0.624 \\ 
& SDL \cite{montenegro2024aim} & 0.055 & 0.059 & 0.168 & 0.051 & 0.027 & 0.056 & 0.099 & 0.095 & 0.640 \\ 
& Glow (AT) & 0.054 & 0.051 & 0.179 & 0.076 & 0.030 & 0.054 & 0.104 & 0.089 & 0.856 \\ 
& Glow (NA) & 0.066 & 0.065 & 0.182 & 0.075 & 0.035 & 0.061 & 0.116 & 0.088 & 0.855 \\ 
 \bottomrule
\end{tabular}
\caption{Results of mean absolute error $\downarrow$ between the asymmetry metrics computed on the template and generated images, and SSIM $\uparrow$ between original and morphed images of the CP dataset. The baseline compares the original pre-operative image with the template (post-operative image).}
\label{tab:final-results}
\end{table}

\begin{figure}[t!]
\centering
\includegraphics[width=1.0\textwidth]{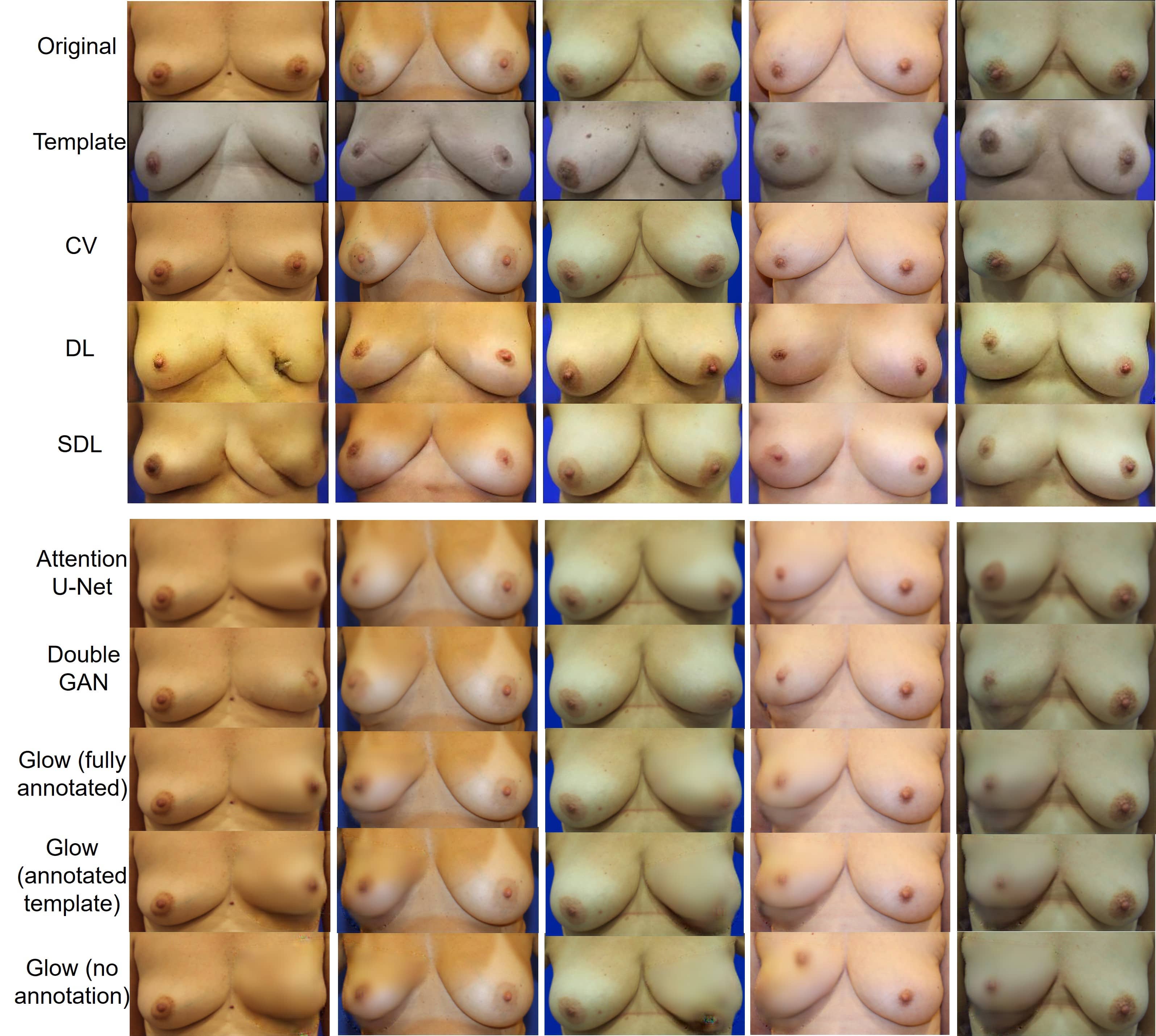}
\caption{Results of transferring asymmetries from post-operative templates into the original pre-operative images on the CP dataset.}
\label{fig:asymmetry-transfer-results}
\end{figure}

The proposed GAN-based and attention-based approaches surpass the state of the art concerning image realism, achieving competitive results in the nipple asymmetry metrics. The conventional approach (CV) proposed in \cite{freitas2024isbi} surpasses our methods regarding the manipulation of the lower breast contour exposed in the LBC metric. Regarding methods that do not require the breast images to be annotated, the proposed Glow-based network surpasses the deep learning model (DL) proposed in \cite{freitas2024isbi} and the simplified model (SDL) proposed in \cite{montenegro2024aim} in image realism, presenting images that closely resemble the original one. Nevertheless, the DL and SDL models generally produce better results than the Glow-based model in the asymmetry metrics. Fig.~\ref{fig:asymmetry-transfer-results} shows examples of results, where it is possible to see that the proposed models produce images that resemble the original while manipulating both breast shape and nipple position. Nevertheless, in images with a more intense manipulation of the lower breast (fourth column), the proposed methods present some artifacts under the manipulated breast. The Glow model without annotations sometimes misplaces the nipple and lower breast due to the accumulated error in the predicted annotations.

\subsection{Ablation Studies}

This section describes various ablation studies used to further validate some implementation choices in the non-invertible and invertible networks. For the non-invertible networks, we evaluate whether the models can preserve color alterations that occur within the covered region and the impact of the shape of the covered region on the input to the inpainting models. For the invertible pipeline, we measure the impact of each step of the post-processing pipeline of the inpainting process.

\subsubsection{Impact of Color Alterations within Covered Breast}

In this experiment, we verify the proposed models' capacity to preserve color alterations within the removed breast region. In specific, we alter the hue of the covered breast on the original image and verify whether the best-performing inpainting models (Double GAN and Attention U-Net) can recognize and preserve these perturbations. 

\begin{figure}[h]
\centering
\includegraphics[width=1.0\textwidth]{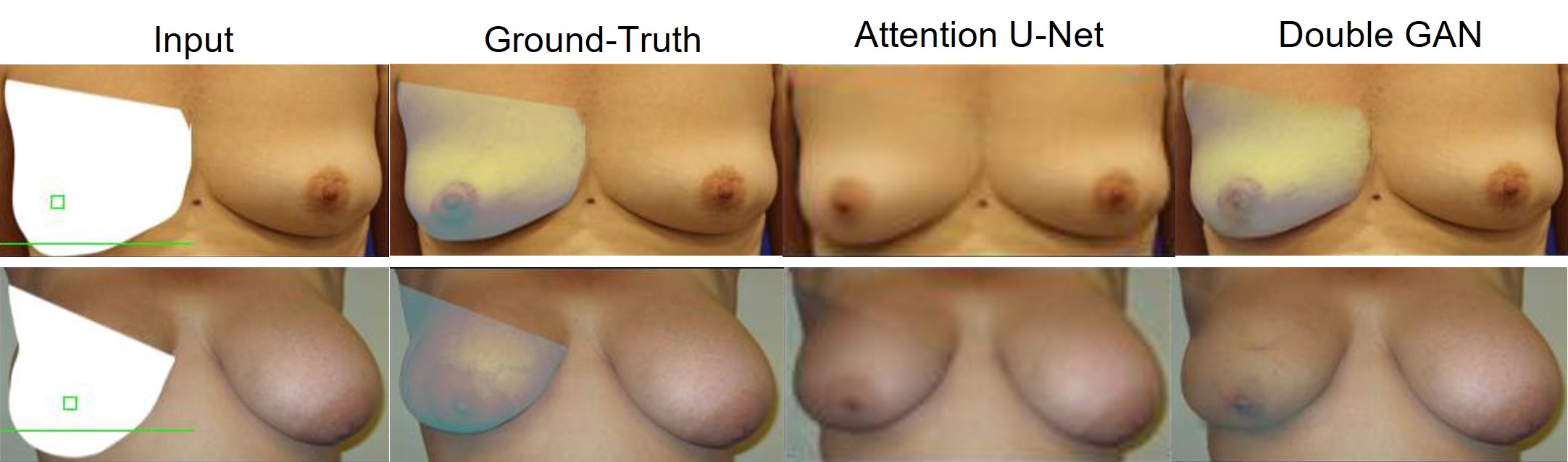}
\caption{Visual results of ablation study  to verify whether the models can reconstruct images in the presence of color alterations within the covered breast.}
\label{fig:color-ablation}
\end{figure}

The results on Table~\ref{tab:ablation-color} suggest that both the models have more difficulties reconstructing the perturbed images than the original ones, which is expected as the perturbations alter the probability distribution of the images and given the difficulties of deep learning models at generalizing to out-of-distribution settings. Nevertheless, the examples displayed in Fig.~\ref{fig:color-ablation} show that the Double GAN model preserves the color alterations that occur within the breast to some extent, unlike the Attention U-Net.

\begin{table}[h]
\centering
\begin{tabular}{l|l|ccc|ccc}
\toprule
\multirow{2}{*}{Model} &\multirow{2}{*}{Perturbation} & \multicolumn{3}{c|}{CP} & \multicolumn{3}{c}{BC} \\
&& SSIM & LPIPS & PSNR & SSIM & LPIPS & PSNR \\ \midrule
A. U-Net & Hue Alteration& 0.910 & 0.128 & 27.44 & 0.920 & 0.109 & 26.86 \\
A. U-Net & None & \textbf{0.950} & \textbf{0.081} & \textbf{32.76} & \textbf{0.966} & \textbf{0.053} & \textbf{34.05} \\
\midrule
D. GAN & Hue Alteration & 0.949 & 0.100 & 32.10 & 0.938 & 0.154 & 27.83 \\
D. GAN & None & \textbf{0.971} & \textbf{0.060} & \textbf{36.29} & \textbf{0.972} & \textbf{0.052} & \textbf{35.94} \\
\bottomrule
\end{tabular}
\caption{Results of the ablation study to measure impact of color alterations in the breast on the inpainting results, highlighting SSIM $\uparrow$, LPIPS $\downarrow$ and PSNR $\uparrow$ between the original image and its reconstruction for each experiment. }
\label{tab:ablation-color}
\end{table}

\subsubsection{Impact of Covered Region on Inpainting}

This experiment aims to measure the impact of the covered region on the inpainting process, verifying whether the inpainting procedure can be applied in the absence of complete breast contour annotations on the original image (with only lower breast and nipple annotations). More specifically, we compare the performance of the best-performing non-invertible networks (Double GAN and Attention U-Net) when applied to images where the covered region is only the target breast and the entire image half that contains the target breast. 

\begin{table}[h]
\centering
\begin{tabular}{l|l|ccc|ccc}
\toprule
\multirow{2}{*}{Model} &\multirow{2}{*}{Covered Area} & \multicolumn{3}{c|}{CP} & \multicolumn{3}{c}{BC} \\
&& SSIM & LPIPS & PSNR & SSIM & LPIPS & PSNR \\ \midrule
A. U-Net & Image Half & 0.896 & 0.140 & 27.05 & 0.880 & 0.150 & 25.36 \\
A. U-Net & Only Breast & \textbf{0.950} & \textbf{0.081} & \textbf{32.76} & \textbf{0.966} & \textbf{0.053} & \textbf{34.05} \\
\midrule
D. GAN & Image Half & 0.952 & 0.084 & 28.23 & 0.940 & 0.085 & 28.95 \\
D. GAN &Only Breast & \textbf{0.971} & \textbf{0.060} & \textbf{36.29} & \textbf{0.972} & \textbf{0.052} & \textbf{35.94} \\
\bottomrule
\end{tabular}
\caption{Results of the ablation study to measure impact of covered region of the input images on the inpainting results, highlighting SSIM $\uparrow$, LPIPS $\downarrow$ and PSNR $\uparrow$ between the original image and its reconstruction for each experiment. }
\label{tab:ablation-mask}
\end{table}

\begin{figure}[h]
\centering
\includegraphics[width=1.0\textwidth]{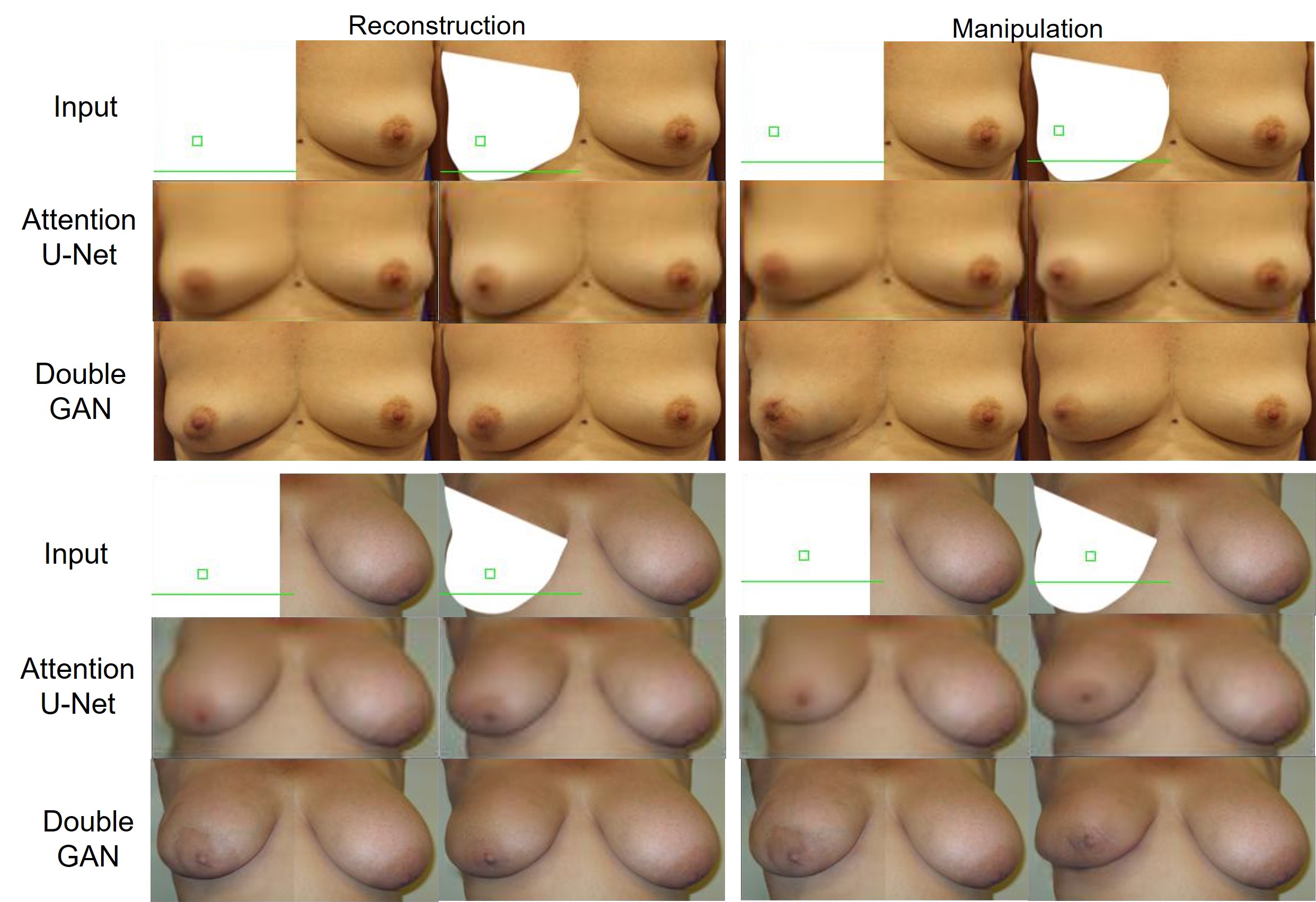}
\caption{Visual results of ablation study  to measure impact of covered region of the input images on the inpainting results.}
\label{fig:mask-ablation}
\end{figure}

The results depicted on Table~\ref{tab:ablation-mask} suggest that covering the entire image half that contains the breast leads to slightly worse visual quality than covering just the breast, which is expected as the area that needs to be inpainted is bigger. Furthermore, the visual results on Fig.~\ref{fig:mask-ablation} suggest that the Double GAN has difficulties manipulating the nipple and breast shape when the entire image half is covered, as it relies more on the information provided by the original breast given as input rather than the masked image. The Attention U-Net succeeds at altering the breast shape and nipple even when half the image is covered but produces blurrier results. These experiments suggest that complete breast contour annotations are required for optimal performance.

\subsubsection{Impact of post-processing steps on Glow-based model}

In this experiment, we measure the impact of each post-processing step on the realism of images generated using the invertible Glow-based U-Net. In specific, we compare four different settings: 
\begin{itemize}
    \item \textbf{Direct Prediction}: apply the inpainting direction directly on the prediction of the segmented image obtained using the segmentation direction of the network, without any post-processing step. Note that in this setting, it is not possible to reposition the nipple.
    \item \textbf{No Resizing}: removing the resizing steps of the post-processing pipeline.
    \item \textbf{No Closing Operation}: removing the closing operation applied to close the gaps left by the prediction of the nipple and lower breast contour positions.
    \item \textbf{Full Post-Processing}: apply all the post-processing steps.
\end{itemize}

Since the goal of these experiments is to verify the impact of the post-processing steps on the realism of the results, we position the nipple and lower breast contour annotations in their ground-truth position. 

\begin{table}[h]
\centering
\begin{tabular}{l|ccc|ccc}
\toprule
\multirow{2}{*}{Method} & \multicolumn{3}{c|}{CP} & \multicolumn{3}{c}{BC} \\
& SSIM & LPIPS & PSNR & SSIM & LPIPS & PSNR \\ \midrule
Direct Prediction & 0.889 & \textbf{0.161} & 25.90 & 0.894 & 0.133 & 24.90 \\
No Resizing & 0.796 & 0.295 & 23.83 & 0.852 & 0.209 & 22.74 \\
No Closing Operation & 0.766 & 0.304 & 22.12 & 0.832 & 0.217 & 21.05 \\
Full Post-Processing & \textbf{0.891} & 0.211 & \textbf{29.29} & \textbf{0.932} & \textbf{0.116} & \textbf{29.98} \\
\bottomrule
\end{tabular}
\caption{Results of the ablation study to measure the impact of the post-processing steps, depicting SSIM $\uparrow$, LPIPS $\downarrow$ and PSNR $\uparrow$ between the original image and its reconstruction. }
\label{tab:ablation-glow}
\end{table}

\begin{figure}[H]
\centering
\includegraphics[width=1.0\textwidth]{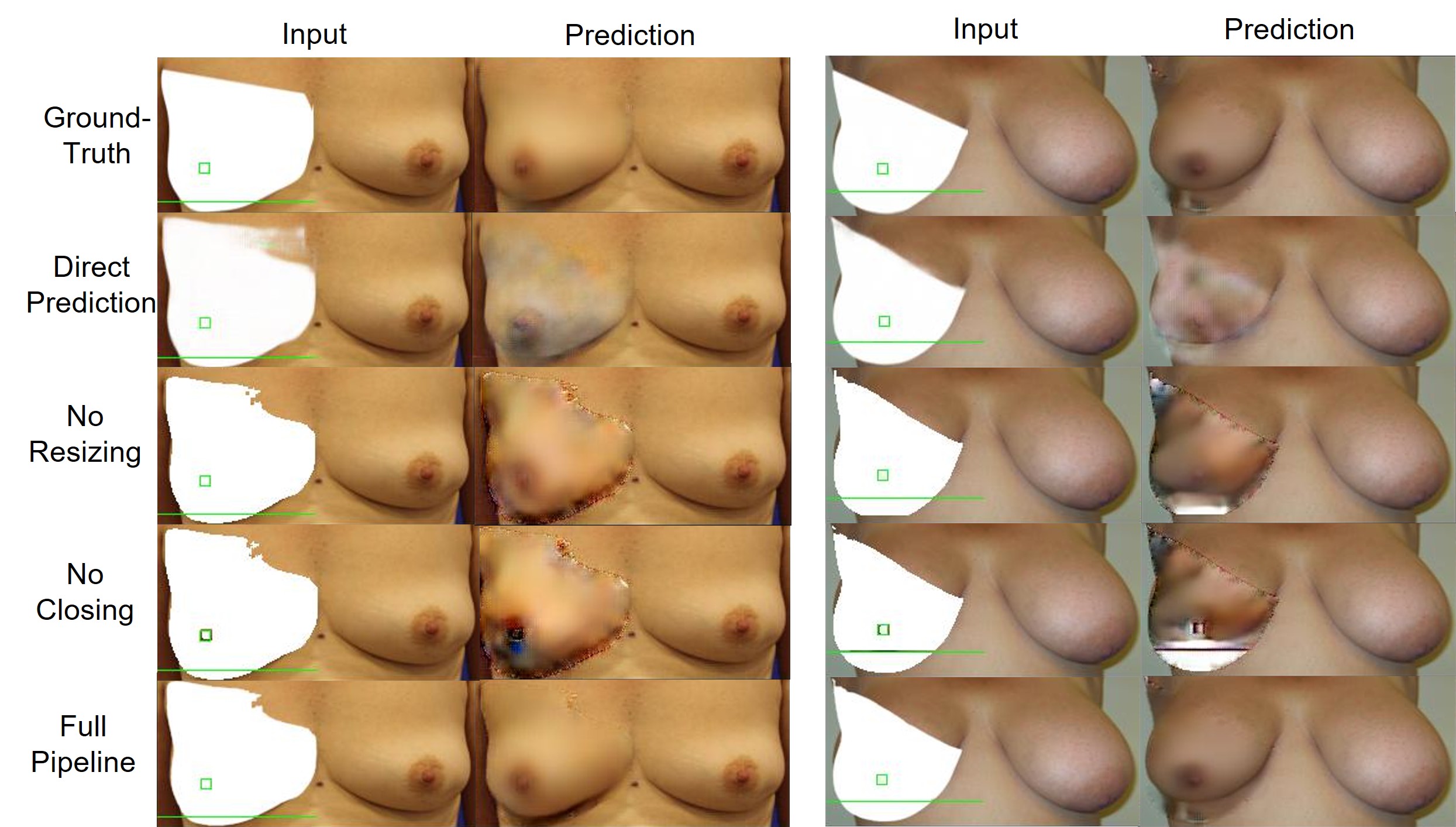}
\caption{Visual results of ablation study defined to measure the impact of each post-processing step on images reconstructed using the Glow-based network.}
\label{fig:glow-ablation}
\end{figure}

The results depicted in Table~\ref{tab:ablation-glow} show that the full post-processing pipeline produces the best results in terms of visual quality. These results are supported by the examples provided in Fig.~\ref{fig:glow-ablation}, which show that removing any step of the post-processing pipeline leads to poor reconstruction of the breast region.

\section{Conclusions}

We proposed inpainting strategies to manipulate asymmetries in breast images, for the purposes of predicting the aesthetic outcomes of breast cancer treatment. We proposed non-invertible inpainting models that surpass the state of the art at generating realistic images and successfully manipulating both breast and nipple asymmetries. Furthermore, we experimented with invertible architectures to enable the manipulation of unlabelled breast images, generating more realistic images than the state of the art unsupervised techniques, while sacrificing some accuracy in the alteration of the nipple. 

Since the models require annotated data during training, we were only able to use small datasets of annotated breast images. As such, future work considers using semi-supervised learning strategies to introduce unlabelled data into the networks' training, aiming to improve the quality of the generated images and the prediction of the nipple and lower breast in the invertible networks. We also consider manipulating other characteristics caused by breast cancer treatment (surgery and radiotherapy), including scars and skin color alterations. 

\section*{Acknowledgments}
This work has received funding from the European Union’s Horizon Europe research and innovation programme under the Grant Agreement 101057389-CINDERELLA project, and from the FCT - Foundation for Science and Technology Portugal, within PhD grant 2022.14516.BD.

\section*{Compliance with Ethical Standards}
This study was performed in line with the principles of the Declaration of Helsinki. Approval was granted by the Ethics Committee of the Champalimaud Foundation (``Conselho de Ética da Fundação Champalimaud'') under the CINDERELLA Project on the 24th of May 2022.

\bibliographystyle{unsrt}
\bibliography{refs}

\end{document}